\documentclass[11pt]{article}

\usepackage[final]{acl}

\usepackage{times}
\usepackage{latexsym}

\usepackage[T1]{fontenc}

\usepackage[utf8]{inputenc}

\usepackage{microtype}

\usepackage{inconsolata}

\usepackage{graphicx}
 \usepackage{amsmath}
\usepackage{changes}
\definechangesauthor[name={Pierre}, color=yellow!85!black]{pim}
\definechangesauthor[name={Fabien}, color=blue]{fab}
\definechangesauthor[name={Catherine}, color=purple]{cat}
\definechangesauthor[name={Hanna}, color=brown]{haa}

\title{SEF-CLGC at SemEval-2026 Task 11: Logical Notation Impact on Language Model Performance}

\author{
  \textbf{Hanna Abi Akl\textsuperscript{1,2}},
  \textbf{Fabien Gandon\textsuperscript{1}},
  \textbf{Catherine Faron\textsuperscript{1}},
  \textbf{Pierre Monnin\textsuperscript{1}} \\
  \textsuperscript{1}Université Côte d’Azur, Inria, CNRS, I3S, Sophia Antipolis, France \\
  \textsuperscript{2}Data ScienceTech Institute, Paris, France \\
\href{mailto:email@domain}{hanna.abi-akl@inria.fr}
}

\begin{document}

\maketitle

\begin{abstract}
This paper revisits our pipeline called Syllogistic Evaluation Framework-Common Logic Grammar Construction (SEF-CLGC). We combine formal logical notations with Small Language Models (SLMs) to evaluate reasoning performance on the SemEval-2026 Task 11 Subtask 1: Disentangling Content and Formal Reasoning in Large Language Models. 
Our experiments show that by relying solely on SLMs, trained on a combination of natural and symbolic languages, our best model achieves a content score of 27.80\% on the task while significantly lowering the content bias in reasoning.   

\end{abstract}

\section{Introduction}
\label{sec:introduction}

Against the trend of scaling bigger language models (LMs), Small Language Models (SLMs) have re-emerged as powerful agents capable of performing complex tasks like reasoning in many domains \cite{masri2026benchmarking, srivastava2025towards}. This emergence has led to benchmarking SLMs across a variety of reasoning tasks like mathematics and common sense problems \cite{zhuang2025technical}. Further exploration has led to investigating neural-based \cite{wang2025slm, kim2025guiding} and neuro-symbolic \cite{lyu-etal-2023-faithful, quan2024verification, han2025enhancing} techniques to enhance their reasoning abilities. The SemEval-2026 Task 11 Subtask 1 \cite{valentino-etal-2026-semeval} challenges LM reasoning on determining the validity of syllogisms. We leverage the Syllogistic Evaluation Framework-Common Logic Grammar Construction (SEF-CLGC) pipeline previously introduced in \cite{akl:hal-05248053} to test the reasoning capabilities of SLMs by training them on different logical notations inspired from formal Knowledge Representation (KR) languages and assessing their performance on the challenge. 

\section{Background}
\label{sec:background}

\subsection{Task Description}
\label{sec:sub_task_description}

The goal of Task 11 Subtask 1 is to determine the correct validity label (i.e. "true" or "false") of a given syllogism. A data point consists of a syllogism composed of premises and a conclusion in natural language (English) and associated information. Table \ref{tab:data_example} is an example. Plausibility indicates whether the arguments of a syllogism are aligned (i.e. "true") or misaligned (i.e. "false") with real-world knowledge. This subtask has 2 phases: a Training phase to prepare and train participating systems on pilot (80 syllogisms) and training (960 syllogisms) sets, and the official Evaluation phase where systems are evaluated and ranked on a blind evaluation set of 191 syllogisms.

\begin{table}
  \centering
  \resizebox{\linewidth}{!}{
  \begin{tabular}{p{0.3\linewidth}  p{0.5\linewidth}llll}
    \hline
    \textbf{ID} & \textbf{Syllogism} & \textbf{Validity} & \textbf{Plausibility} \\
    \hline
50146f21-d265-4e3a-8d93-8165cdbe89a3 & All cars are a type of vehicle. No animal is a car. Therefore, no animal can be a vehicle. & False & True \\
\hline
08408587-3887-4246-9d6f-7a4492ad48c7 & Anyone who is a rose is red. Some flowers are not red. From this, all flowers are roses. & False & False \\
\hline
  \end{tabular}
  }
  \caption{Example data.}
  \label{tab:data_example}
\end{table}

\subsection{Related Work}
\label{sec:sub_related_work}

Different systems have been proposed in the literature to predict the validity of a syllogism. Works leveraging small and large LMs \cite{eisape2024systematic, ozeki2024exploring} show that larger models make less mistakes but are still prone to the same reasoning biases (e.g. syllogistic fallacies) as humans. Other works \cite{dasgupta2022language, bertolazzi2024systematic} compare Supervised Fine-Tuning (SFT) and In-Context Learning (ICL) strategies and show that SFT has a better mitigating effect than ICL on bias on small and mid-sized LMs. Reasoning limitations in LMs paved the way for neuro-symbolic systems that integrate rules \cite{seals2024evaluating, valentino2025mitigating, wysocka2025syllobio} in the model prompt to control bias and enhance performance. Other areas of research leverage prompting techniques like Chain-Of-Thought (COT) and explanation generation as a method of auto-correction and auto-evaluation of LM reasoning \cite{xu-etal-2024-faithful}. Lastly, multi-stage systems that translate syllogisms from Natural Language (NL) to formal languages like First-Order Logic (FOL) and combine natural explanations with theorem-proving \cite{ranaldi2025improving, kim-etal-2025-reasoning, maraia2026abstract} show boosted performance in predicting syllogism validity. Our work follows in that direction.

\section{System Architecture}
\label{sec:system_architecture}

\subsection{Enhanced SEF-CLGC}
\label{sec:enhanced-sef-clgc}

We reuse the SEF-CLGC pipeline from \cite{akl:hal-05248053} which transforms syllogisms in FOL notation into other logical notations: CLIF, CGIF and TFL+. We enhance the pipeline by introducing two new notations: CLINGO and a custom MINIFOL2.

\paragraph{CLINGO}\label{sec:sub_clingo} \cite{gebser2014clingo} is a language for Answer Set Programming (ASP), a form of declarative logic programming used to model and solve combinatorial search problems.

\paragraph{MINIFOL2}\label{sec:sub_minifol} spans from the custom MINIFOL notation introduced in \cite{akl:hal-05248053} which replaces FOL operators with their Boolean equivalents (e.g. "$\wedge$" with “\&”).  MINIFOL2 also eliminates the "\textit{$\exists$}" quantifier from the syntax.  The notation is used as a baseline to study the effects of slight syntactic changes on LMs.

\subsection{Pipeline}
\label{sec:sub_pipeline}
\begin{figure*}[t]
  \includegraphics[width=1.0\linewidth]{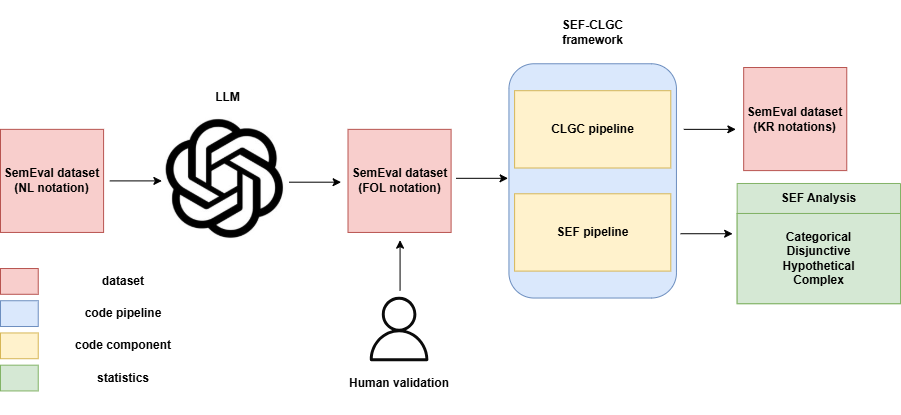} \hfill
  \includegraphics[width=0\linewidth]{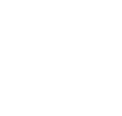}
  \caption {Dataset generation workflow.}
  \label{fig:workflow}
\end{figure*}

The methodology pipeline is shown in Figure \ref{fig:workflow}. 

\paragraph{NL-FOL Translation:}\label{sec:sub_sub_nl_fol_translation_pipeline} The starting point for SEF-CLGC is FOL notation. Since the Subtask data is provided in NL, we first translate the syllogisms into FOL notation using OpenAI's ChatGPT 5.2 model\footnote{\url{https://chatgpt.com/share/699c606c-38c0-800e-ac41-6a55c246dd57}}. We chose this model because preliminary tests showed good performance but other models will be tested in extensions of this work. Translations are validated manually on a random 20\% sample from the training set due to its size as well as the entire evaluation set from the evaluation phase.

\paragraph{SEF-CLGC:}\label{sec:sub_sub_sef_clgc_pipeline} The dataset in FOL notation is then given to the SEF-CLGC framework. The SEF component categorizes syllogisms into 4 categories as per \cite{akl:hal-05248053}: Hypothetical (i.e. containing an implication), Disjunctive (i.e. containing a disjunction), Categorical (i.e. any syllogism of 2 premises and a conclusion not belonging to the aforementioned categories) and Complex (i.e. not belonging to any other category). Appendix \ref{sec:appendix_sef} shows examples of SEF categories. The CLGC component uses the Backus-Naur Form (BNF) grammar of logical notations including FOL to generate the Abstract Syntax Tree (AST) of each syllogism and validate it using a syntactic parser. Each syllogism is then transformed to the target logical notation from the AST and the BNF grammar of that notation. The resulting dataset contains all initial syllogisms along with their transformations in each of the logical notations (i.e. KR notations in Figure \ref{fig:workflow}). Appendix \ref{sec:appendix_sef} also shows examples of the transformation from FOL to logical notations.

\paragraph{Validity Prediction:}\label{sec:sub_sub_validity_prediction} Syllogisms in a notation (e.g. NL, FOL) or a combination of notations (e.g. NL-FOL, FOL-CLIF-CGIF) are given with their validity labels as input in SFT to LMs for training and evaluation during the Training phase. For the blind Evaluation, only the syllogisms are given to the LMs to predict the validity labels. Based on previous work \cite{han2024folio, akl:hal-05248053}, we limit ourselves to very small LMs (i.e. less than 1 billion parameters) for their frugality and good results on this type of task.

\section{Experimental Setup}
\label{sec:experiments}

We combine the official task pilot and training datasets as our working set and split it into train/validation/test sets to fine-tune our models. These splits are frozen and used for all our experiments.
We use Google's Flan-T5-small\footnote{\url{https://huggingface.co/google/flan-t5-small}} and large\footnote{\url{https://huggingface.co/google/flan-t5-large}} and fine-tune them on Google Cloud T4 and A100 GPUs respectively. Inference is performed on an A100 GPU.

\subsection{SEMEVAL Models}
\label{sec:models}

\paragraph{SEMEVAL Models}\label{sec:sub_semeval_models} are 
all vanilla Flan-T5 models directly trained on the working dataset splits. All models are trained on 5 epochs with a learning rate of $10^{-5}$ and a batch size of 4. All other parameters are set to default.

\paragraph{FOLIO-SEMEVAL Models}\label{sec:sub_folio_semeval_models} are Flan-T5 models that have already been fine-tuned on the FOLIO dataset as per \cite{akl:hal-05248053} with the same parameters as SEMEVAL Models for epochs, learning rate and batch size and the rest being default. The models are then also fine-tuned on the working dataset, again under the same conditions as the SEMEVAL models.

\subsection{Evaluation}
\label{sec:sub_evaluation}

All models receive the pair (syllogism, validity label) in one or more notation as input. Evaluation is based on the Content Score (CS): 
\begin{equation}
  CS = \frac{ACC}{1+\log(1+CE)}\
  \label{eq:eval_metric}
\end{equation}
\noindent where ACC is the overall accuracy and CE is the Content Effect. Since a syllogism can have one of two validity values (i.e. Valid or Invalid) for each of its two plausibility values (i.e. Plausible or Implausible), the CE is the average accuracy difference between Plausible (i.e. Plausible\_Valid + Plausible\_Invalid) and Implausible (i.e. Implausible\_Valid + Implausible\_Invalid) syllogisms. 

\section{Results}
\label{sec:results}

We present our training and evaluation results for both SEMEVAL and FOLIO-SEMEVAL models. Training results are scored on Accuracy only since the CE and CS calculations were performed on the official evaluation set. 

\subsection{SEMEVAL Models}
\label{sec:sub_semeval_results}

Table \ref{tab:semeval_train_results} shows the training results for the SEMEVAL models. 
NL Flan-T5-small is used as a baseline performance for very small language models. Overall, of the Flan-T5-large models, NL performs best as it is the most seen notation for these models, followed closely by the NL-FOL and NL-CLIF notations. 
Adding more notations to the input does not seem to boost performance as shown by NL-FOL-CLIF which fails to beat NL-FOL or NL-CLIF. Table \ref{tab:semeval_eval_results} shows the official results for this family. NL retains top spot, followed by NL-FOL since these two notations are the most widely seen in the pre-training of these models among the notations used for the task. NL-CLIF is bested by NL-FOL-CLIF possibly boosted by FOL and NL-CLINGO since CLINGO is closer to FOL notation.
MINIFOL2 performs badly in both cases since it combines FOL syntax with Boolean operators that LMs are not accustomed to seeing together which may explain why it breaks down. TFL+ proves to be too abstract and cannot be learned efficiently.

\begin{table}
  \centering
  \begin{tabular}{lllll}
    \hline
    \textbf{Notation} & \textbf{Acc} & \textbf{Pr} & \textbf{Re} & \textbf{F1} \\
    \hline
\textbf{NL} & \textbf{0.92}	& \textbf{0.92}	& \textbf{0.92}	& \textbf{0.92} \\
\underline{NL-FOL} & \underline{0.91}	& \underline{0.91}	& \underline{0.91}	& \underline{0.91} \\
NL-CLIF & 0.88	& 0.88	& 0.88	& 0.88 \\
NL-FOL-CLIF & 0.88	& 0.89	& 0.88	& 0.88 \\
NL-CLINGO & 0.87	& 0.87	& 0.87	& 0.87 \\
FOL & 0.80	& 0.81	& 0.80	& 0.80 \\
$^*$NL & 0.78	& 0.79	& 0.78	& 0.78 \\
CLINGO & 0.77	& 0.81	& 0.77	& 0.77 \\
CGIF & 0.75	& 0.76	& 0.75	& 0.75 \\
CLIF & 0.74	& 0.75	& 0.74	& 0.73 \\
MINIFOL2 & 0.72	& 0.72	& 0.72	& 0.71 \\
FOL-CLIF-CLINGO & 0.70	& 0.72	& 0.70	& 0.69 \\
TFL+ & 0.61	& 0.61	& 0.61	& 0.61 \\
\hline
  \end{tabular}
  \caption{SEMEVAL Flan-T5-large training results: best results in bold and second-best underlined. Acc = Accuracy; Pr = Precision; Re = Recall; F1 = Weighted F1. $^*$ Flan-T5-small is used here as a baseline.}
  \label{tab:semeval_train_results}
\end{table}

\begin{table}
  \centering
  \begin{tabular}{llll}
    \hline
    \textbf{Notation} & \textbf{Acc} & \textbf{CE} & \textbf{CS} \\
    \hline
\textbf{NL} & \textbf{90.05}	& \textbf{9.57}	& \textbf{26.81} \\
\underline{NL-FOL} & \underline{89.00}	& \underline{10.68}	& \underline{25.73} \\
NL-FOL-CLIF & 84.29	& \underline{10.68}	& 24.37 \\
NL-CLINGO & 84.29	& 10.70	& 24.36 \\
NL-CLIF & 83.76	& 10.70	& 24.21 \\
CLINGO & 67.53	& 23.95	& 16.01 \\
CGIF & 66.49	& 28.12	& 15.21 \\
CLIF & 65.96	& 29.16	& 14.96 \\
MINIFOL2 & 64.92	& 23.95	& 15.39 \\
FOL-CLIF-CLINGO & 64.39	& 27.08	& 14.85 \\
FOL & 61.25	& 31.25	& 13.69 \\
TFL+ & 59.16	& 5.89	& 20.18 \\
\hline
  \end{tabular}
  \caption{SEMEVAL Flan-T5-large evaluation results: best results in bold and second-best underlined. Acc = Accuracy; CE = Content Effect; CS = Combined Score.}
  \label{tab:semeval_eval_results}
\end{table}

\subsection{FOLIO-SEMEVAL Models}
\label{sec:sub_folio_semeval_results}

\begin{table}
  \centering
  \begin{tabular}{lllll}
    \hline
    \textbf{Notation} & \textbf{Acc} & \textbf{Pr} & \textbf{Re} & \textbf{F1} \\
    \hline
\textbf{NL-CLIF} & \textbf{0.95} & \textbf{0.95} & \textbf{0.95} & \textbf{0.95} \\
\underline{NL} & \underline{0.93} &	\underline{0.93} &	\underline{0.93} &	\underline{0.93} \\
NL-FOL & 0.92	& 0.92	& 0.92	& 0.92 \\
CLIF & 0.85	& 0.85	& 0.85	& 0.85 \\
CLINGO & 0.84	& 0.84	& 0.84	& 0.84 \\
FOL & 0.81	& 0.81	& 0.81	& 0.81 \\
CGIF & 0.75	& 0.76	& 0.75	& 0.75 \\
MINIFOL2 & 0.75	& 0.75	& 0.75	& 0.74 \\
TFL+ & 0.55	& 0.56	& 0.55	& 0.54 \\
\hline
  \end{tabular}
  \caption{FOLIO-SEMEVAL Flan-T5-large training results: best results in bold and second-best underlined.}
  \label{tab:folio_train_results}
\end{table}

\begin{table}
  \centering
  \begin{tabular}{llll}
    \hline
    \textbf{Notation} & \textbf{Acc} & \textbf{CE} & \textbf{CS} \\
    \hline
NL-FOL & \underline{90.57}	& \underline{8.55}	& \textbf{27.80} \\
NL & \textbf{93.19}	& 9.57	& \underline{27.74} \\
NL-CLIF & 89.00	& 13.85	& 24.06 \\
CLIF & 80.00	& 50.00	& 16.22 \\
CLINGO & 74.34	& 16.66	& 19.20 \\
CGIF & 69.10	& 20.83	& 16.92 \\
FOL & 66.49	& \textbf{3.50}	& 26.54 \\
TFL+ & 58.11	& 10.41	& 16.91 \\
$^*$MINIFOL2 & N/A	& N/A	& N/A \\
\hline
  \end{tabular}
  \caption{FOLIO-SEMEVAL Flan-T5-large evaluation results: best results in bold and second-best underlined. $^*$ scoring timed out on the evaluation platform.}
  \label{tab:folio_eval_results}
\end{table}

Table \ref{tab:folio_train_results} shows the training results for the FOLIO-SEMEVAL models. While NL-CLIF could not beat NL in the SEMEVAL models, its FOLIO-SEMEVAL counterpart having already been fine-tuned on FOLIO outperforms it and even beats NL.
This suggests that the model can learn CLIF well, making it a reliable asset to solve syllogistic problems and underlining the added value of enriching NL with a formal notation. This is further emphasized by the results of the CLIF and CLINGO models which perform very well having only been fine-tuned on these notations on the relatively small FOLIO dataset. NL-CLIF and NL-FOL show the importance of neuro-symbolic integration to augment logical performance for LMs. Table \ref{tab:folio_eval_results} shows a slight change in ranking while retaining the behavior observed in training: NL-FOL takes top spot as the combination of FOL with NL reduces CE and results in a better overall CS score than plain NL. NL-CLIF falls to third place but still performs decently. Possible explanations for the performance loss might be LM NL-FOL translation errors that propagate to other logical notations or the notable increase in CE observed for this particular notation which suggests it learns some forms of plausible syllogisms better than others. Notations like TFL+ and MINIFOL2 perform poorly due to the unfamiliarity of LMs with their syntax which can result in performance breakdown. Prior work shows that re-training LM tokenizers on these syntaxes slightly boosts performance at small scale but breaks down as models scale \cite{akl:hal-05248053}. It is also worth noting that more abstract notations (e.g. TFL+) significantly decrease CE at the expense of reduced accuracy, as do neuro-symbolic combinations of NL and formal notations (e.g. NL-FOL-CLIF). The results also clearly highlight the positive effect of SFT on prior logical datasets to boost logical reasoning in LMs.

\subsection{Ranking}
\label{sec:sub_ranking}

Our overall best FOLIO-SEMEVAL evaluation model (i.e. NL-FOL) ranks $10^{th}$ in Accuracy and $7^{th}$ in CE in Subtask 1. Considering we limited our experiments to SLMs, our results show that it is possible to have a model that is both competitive and frugal. For reproducibility purposes, we openly released the weights of our best training model NL-CLIF\footnote{\url{https://huggingface.co/HannaAbiAkl/LOGIC-NL-CLIF-Flan-T5-Large}} and evaluation model NL\footnote{\url{https://huggingface.co/HannaAbiAkl/LOGIC-NL-Flan-T5-Large}} (best Accuracy) and will soon release those for NL-FOL (best CS).

\subsection{Analysis}
\label{sec:sub_analysis}

\subsubsection{Dataset}
\label{sec:sub_dataset}

Appendix \ref{sec:appendix_dataset_stats} shows the SEF classes for the working and evaluation sets. Categorical syllogisms dominate which might explain the performance of the SEMEVAL models since these syllogisms share the same syntactic patterns. Conversely, it might also explain the CE drift in the FOLIO-SEMEVAL models that were pre-fine-tuned on a more diverse dataset of SEF classes. The results suggest further investigation is needed on a dataset with more balanced SEF classes to re-assess the performance of the two families.

\subsubsection{Prediction Errors}
\label{sec:sub_prediction_errors}

In Appendix \ref{sec:appendix_quantitative_errors}, Figure \ref{fig:semeval_validity_plausibility_error_analysis} shows that the best 3 training and evaluation SEMEVAL models (i.e. NL, NL-CLIF and NL-FOL) have a high False Positive (FP) count and a lower False Negative (FN) count. In comparison, Figure \ref{fig:folio_semeval_validity_plausibility_error_analysis} shows that for the same best FOLIO-SEMEVAL models the gap between FP and FN is wider as these models have very high FP and very low FN counts. This suggests that our models are very good at detecting valid syllogisms but prone to making mistakes when predicting invalid ones.

Future analysis could focus on evaluating the decision threshold of these models or the syntactic structure of invalid syllogisms to boost their performance. The greater difference between FP and FN for FOLIO-SEMEVAL models might also be due to their prior exposure to the FOLIO dataset composed of 460 valid versus 351 invalid syllogism which may create the observed bias.

For plausibility, all FOLIO-SEMEVAL models make more mistakes on plausible syllogisms and very few errors on implausible ones from Figure \ref{fig:folio_semeval_validity_plausibility_error_analysis} in Appendix \ref{sec:appendix_quantitative_errors}. Implausible syllogisms are logically structured arguments whose premises are unlikely or contrary to common sense. 
This suggests that these models are capable of handling imagined reasoning scenarios that stray from the common world knowledge they have learned in pre-training. 
The same pattern is observed for SEMEVAL models in Figure \ref{fig:semeval_validity_plausibility_error_analysis} with a reduced difference between plausible and implausible errors. The errors made on plausible syllogisms may be due to their arguments being borrowed from other syllogisms resulting in unfamiliar constructions to the LMs and bad predictions. We cannot attribute with certainty the difference in plausibility errors between the 2 families of models to the FOLIO dataset as the information on the plausibility of syllogisms is not explicitly divulged but offers us another avenue to explore in future research.

Figures \ref{fig:semeval_common_error_analysis} and \ref{fig:folio_semeval_common_error_analysis} in \ref{sec:appendix_qualitative_errors} show the common FP and FN errors for the SEMEVAL and FOLIO-SEMEVAL models respectively. Comparing both figures shows that for FP the highest error count is the common error among all 3 models which suggests that they are likely to reason on invalid syllogisms in similar fashion. The same cannot be said for FN errors which suggests models are very sensitive to changes in input combinations when reasoning on valid syllogisms.
The comparison also shows that for both families NL-CLIF has the highest counts of uncommon errors with the other models in most cases (11.1\%, 41.2\%, 29.4\% and 33.3\% respectively) suggesting that models with this combination of notations are more prone to different reasoning behaviors.

\section{Conclusion}
\label{sec:conclusion}

In this paper, we extended the SEF-CLFC pipeline to the SemEval-2026 Task 11 Subtask 1 dataset. 
Our results show that by combining NL and a logical notation, SLMs can achieve up to 90\% accuracy in classifying syllogism validity on the evaluation set. 
Our method also demonstrates that expressing logical problems in formal notations can reduce content bias. The SemEval challenge opened new questions for us to explore in future work, most notably enriching under-represented SEF categories as well as studying the impact of including more implausible syllogisms on logical notations for SLM reasoning. Possible future work will also include performance comparisons against larger LMs.

\section{Limitations}
\label{sec:limitations}

A limitation of our work in the context of this task is the reliance on a commercial model to translate from NL to FOL. Updates or changes on the model usage could affect the quality of translation and potentially alter subsequent results in the SEF-CLGC pipeline.

\nocite{wysocka2025syllobio}

\bibliography{custom}

\appendix

\section{Appendix}
\label{sec:appendix}

\subsection{Dataset Examples}
\label{sec:appendix_sef}

In this section, we provide a concrete example from the working dataset and the resulting transformations from the SEF-CLGC framework. Table \ref{tab:sef_example} shows the SEF classification on examples from the working dataset. Table \ref{tab:sef_clgc_pipeline_example} shows the resulting CLGC transformation into different logical notations.

\begin{table}
  \centering
  \resizebox{\linewidth}{!}{
  \begin{tabular}{p{0.3\linewidth}  p{0.4\linewidth}lllll}
    \hline
    \textbf{ID} & \textbf{Syllogism} & \textbf{Validity} & \textbf{Plausibility} & \textbf{SEF} \\
    \hline
951df8bb-e9dc-4272-9db7-92fe5d28d337 & Anything that is a dog has fur. There are some poodles that are dogs. There are no poodles that do not have fur. & False & True & Categorical \\
\hline
4480e5d5-495a-4928-a420-a3c74b9268a9 & Every single mammal is an animal. Each and every feline is an animal. This makes it true that every feline is a mammal. & False & True & Hypothetical \\
\hline
  \end{tabular}
  }
  \caption{Example data with SEF categories.}
  \label{tab:sef_example}
\end{table}

\begin{table*}
  \centering
  \resizebox{\linewidth}{!}{
  \begin{tabular}{p{0.2\linewidth} | p{0.4\linewidth} | p{0.2\linewidth} | p{0.2\linewidth} | p{0.2\linewidth} | p{0.2\linewidth} | p{0.2\linewidth}lllllll}
    \hline
    \textbf{ID}  & \textbf{FOL} & \textbf{CLIF} & \textbf{CGIF} & \textbf{CLINGO} & \textbf{TFL+} & \textbf{MINIFOL2} \\
    \hline
951df8bb-e9dc-4272-9db7-92fe5d28d337 & $\forall$x (AnythingThat(x) → DogHasFur(x)) 
$\exists$x (PoodleThat(x) $\wedge$ Dog(x))  
$\forall$x (There(x) → NoPoodleThatDoNotHaveFur(x))
& forall x (anythingthat(x) implies doghasfur(x))
 exists x (poodlethat(x) and dog(x))
 forall x (there(x) implies nopoodlethatdonothavefur(x))
& [@every *x [(anythingthat[(?x)]  doghasfur[(?x)])]
 *x [(poodlethat[(?x)]  dog[(?x)])]
 @every *x [(there[(?x)]  nopoodlethatdonothavefur[(?x)])]
] & forall (anythingthat(x) -: doghasfur(x))
  (poodlethat(x) , dog(x))
 forall (there(x) -: nopoodlethatdonothavefur(x))
& -(+A0-+D0)+(+P1++D1)-(+T0-+N0) & all:x (anythingthat(x) :- doghasfur(x))
 x (poodlethat(x) \& dog(x))
 all:x (there(x) :- nopoodlethatdonothavefur(x)) \\
\hline
4480e5d5-495a-4928-a420-a3c74b9268a9 &  $\forall$x (SingleMammal(x) → Animal(x))
 $\forall$x (Feline(x) → Animal(x))
 $\forall$x (ThiMakeItTrueThatEveryFeline(x) → Mammal(x))
& forall x (singlemammal(x) implies animal(x))
 forall x (feline(x) implies animal(x))
 forall x (thimakeittruethateveryfeline(x) implies mammal(x))
& [@every *x [(singlemammal[(?x)]  animal[(?x)])]
 @every *x [(feline[(?x)]  animal[(?x)])]
 @every *x [(thimakeittruethateveryfeline[(?x)]  mammal[(?x)])]
]& forall (singlemammal(x) -: animal(x))
 forall (feline(x) -: animal(x))
 forall (thimakeittruethateveryfeline(x) -: mammal(x))
& -(+S0-+A0)-(+F0-+A0)-(+T0-+M0) & all:x (singlemammal(x) :- animal(x))
 all:x (feline(x) :- animal(x))
 all:x (thimakeittruethateveryfeline(x) :- mammal(x)) \\
\hline
  \end{tabular}
  }
  \caption{CLGC transformation example on the working dataset.}
  \label{tab:sef_clgc_pipeline_example}
\end{table*}

\subsection{Dataset Statistics}
\label{sec:appendix_dataset_stats}

Table \ref{tab:sef_results} shows the SEF classification statistics on the working dataset splits and the evaluation set. Tables \ref{tab:intra_validity_plausibility} and \ref{tab:inter_validity_plausibility} show the validity and plausibility count distribution on the same sets.

\begin{table}
  \centering
  \begin{tabular}{lllllll}
    \hline
    \textbf{Set} & \textbf{Split} & \textbf{Size} & \textbf{Ca} & \textbf{Hy} & \textbf{Di} & \textbf{Co} \\
    \hline
Working & Train & 561	& \textbf{547} & 14	& 0 & 0 \\
Working & Val & 375 & \textbf{368}	& 7 & 0 & 0 \\
Working & Test & 104	& \textbf{101} & 3 & 0 & 0 \\
Evaluation & Eval & 191	& \textbf{189} & 2 & 0 & 0 \\
\hline
  \end{tabular}
  \caption{SEF classification results on the working and evaluation sets: majority syllogism class in bold. Ca = Categorical; Hy = Hypothetical; Di = Disjunctive; Co = Complex.}
  \label{tab:sef_results}
\end{table}

\begin{table}
  \centering
  \begin{tabular}{lllll}
    \hline
    \textbf{Dataset} & \textbf{V} & \textbf{IV} & \textbf{P} & \textbf{IP} \\
    \hline
Pilot & 40 & 40 & 40 & 40 \\
Training & 480 & 480 & 474 & 486 \\
Evaluation & 96 & 95 & 95 & 96 \\
\hline
  \end{tabular}
  \caption{Task dataset validity and plausibility intra-distribution. V = Valid; IV = Invalid; P = Plausible; IP = Implausible.}
  \label{tab:intra_validity_plausibility}
\end{table}

\begin{table}
  \centering
  \begin{tabular}{lllll}
    \hline
    \textbf{Dataset} & \textbf{V-P} & \textbf{V-I} & \textbf{I-P} & \textbf{I-I} \\
    \hline
Pilot & 20 & 20 & 20 & 20 \\
Training & 240 & 240 & 234 & 246 \\
Evaluation & 48 & 48 & 47 & 48 \\
\hline
  \end{tabular}
  \caption{Task dataset validity and plausibility inter-distribution. V-P = Valid-Plausible; V-I = Valid-Implausible; I-P = Invalid-Plausible; I-I = Invalid-Implausible.}
  \label{tab:inter_validity_plausibility}
\end{table}

\subsection{Error Analysis}
\label{sec:appendix_error_analysis}

\subsubsection{Quantitative Error Analysis}
\label{sec:appendix_quantitative_errors}

\begin{figure*}[t]
  \includegraphics[width=0.5\linewidth]{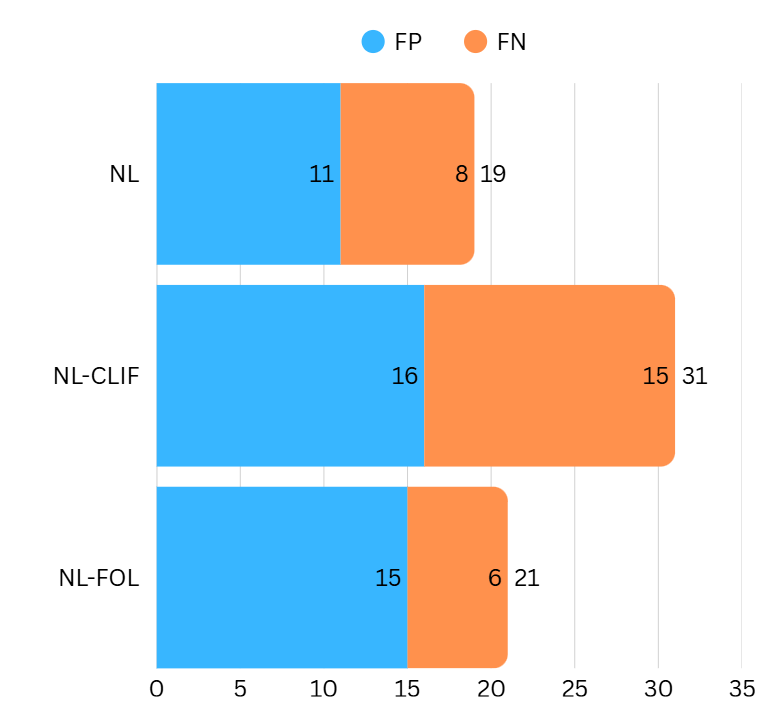}
  \includegraphics[width=0.5\linewidth]{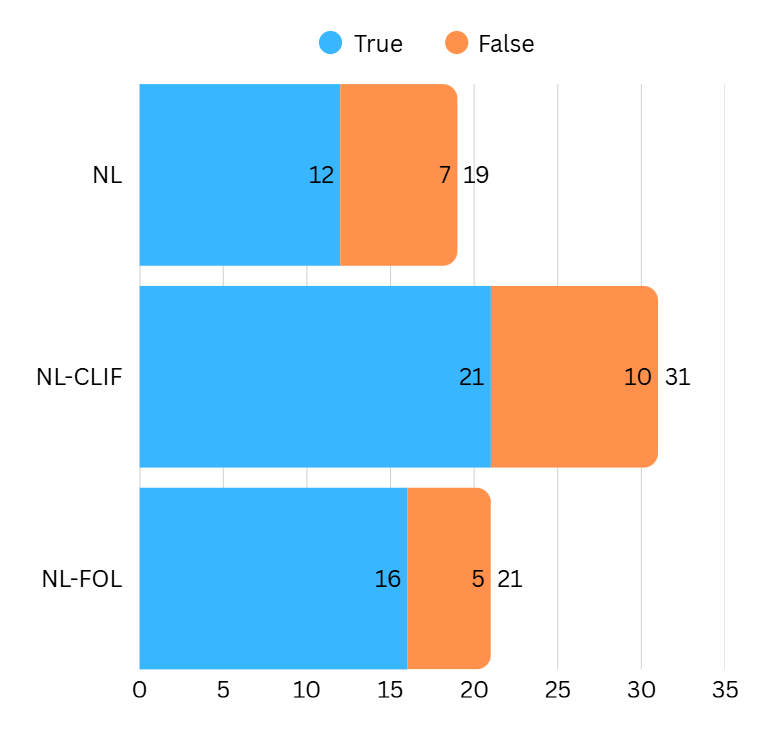}
  \caption {Left: Error analysis of validity prediction of the best SEMEVAL Flan-T5-large model notations. FP = False Positives, FN = False Negatives. Right: Plausibility ground truth of prediction errors for the best SEMEVAL Flan-T5-large models.}
  \label{fig:semeval_validity_plausibility_error_analysis}
\end{figure*}

\begin{figure*}[t]
  \includegraphics[width=0.5\linewidth]{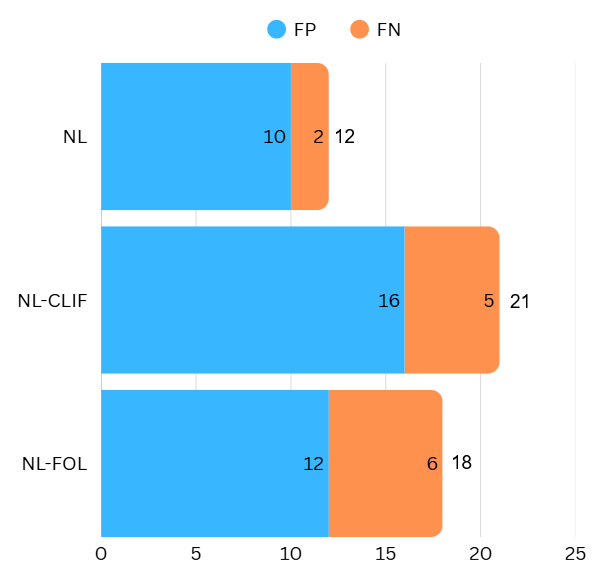}
  \includegraphics[width=0.5\linewidth]{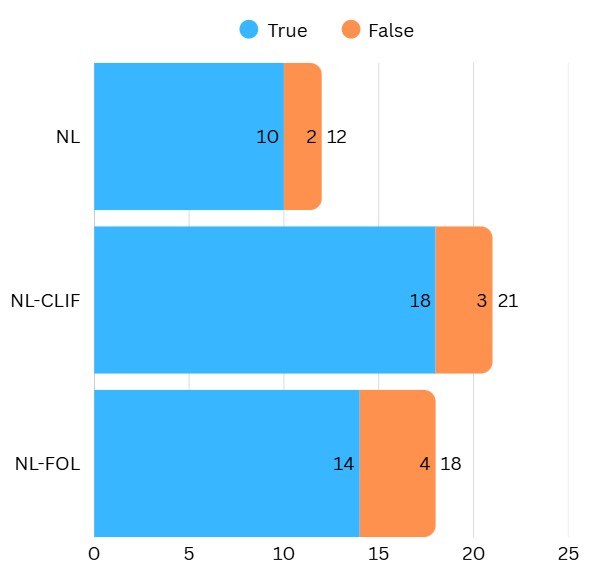}
  \caption {Left: Error analysis of validity prediction of the best FOLIO-SEMEVAL Flan-T5-large model notations. FP = False Positives, FN = False Negatives. Right: Plausibility ground truth of prediction errors for the best FOLIO-SEMEVAL Flan-T5-large models.}
  \label{fig:folio_semeval_validity_plausibility_error_analysis}
\end{figure*}

Figure \ref{fig:semeval_validity_plausibility_error_analysis} shows the FP and FN error distribution on the validity prediction for the 3 best SEMEVAL models as well as the plausibility of the respective syllogisms. Figure \ref{fig:folio_semeval_validity_plausibility_error_analysis} shows the same analysis for the best 3 FOLIO-SEMEVAL models.

\subsubsection{Qualitative Error Analysis}
\label{sec:appendix_qualitative_errors}

\begin{figure*}[t]
  \includegraphics[width=0.5\linewidth]{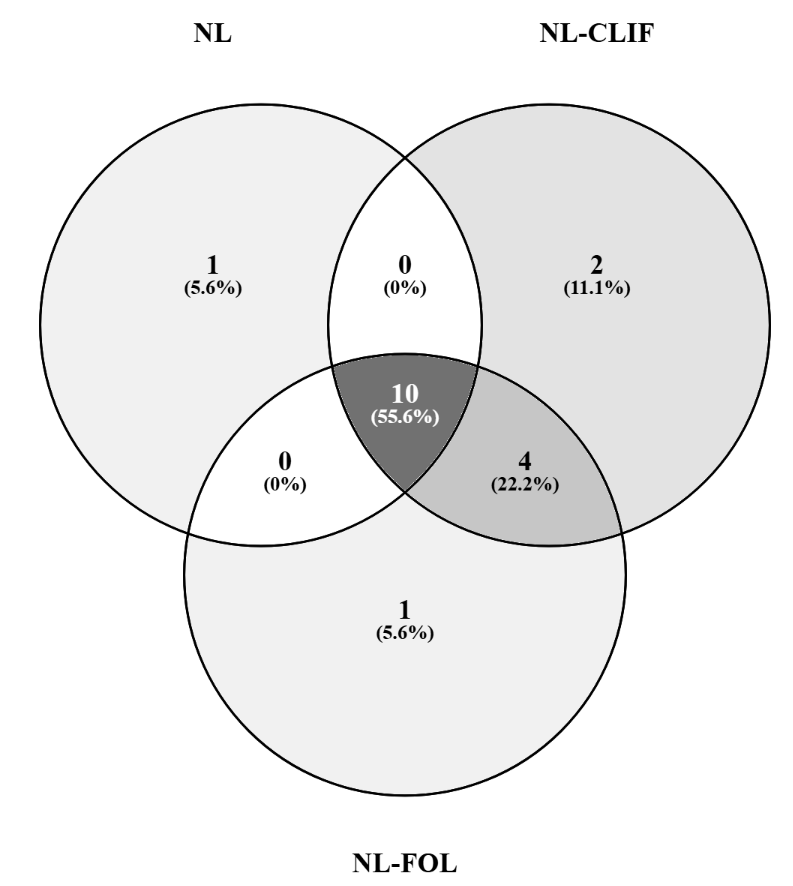}
  \includegraphics[width=0.5\linewidth]{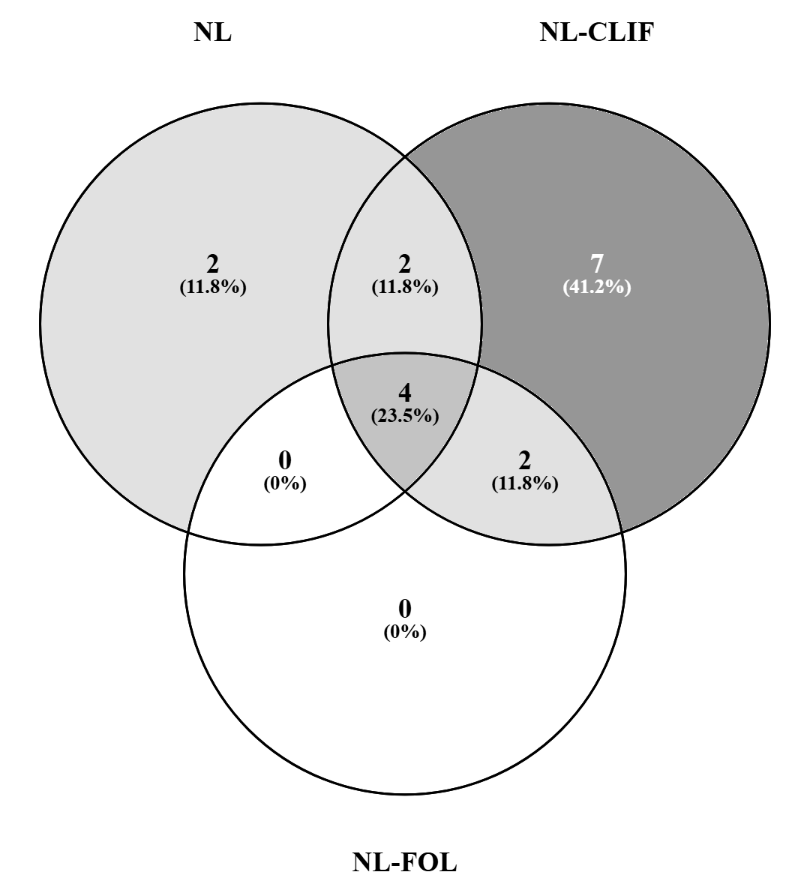}
  \caption {Left: False Positive common errors of the best SEMEVAL Flan-T5-large model notations. Right: False Negative common errors of the best SEMEVAL Flan-T5-large models.}
  \label{fig:semeval_common_error_analysis}
\end{figure*}

\begin{figure*}[t]
  \includegraphics[width=0.5\linewidth]{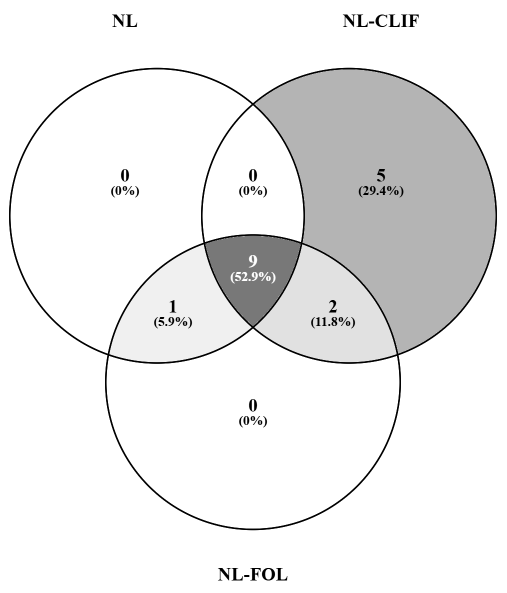}
  \includegraphics[width=0.5\linewidth]{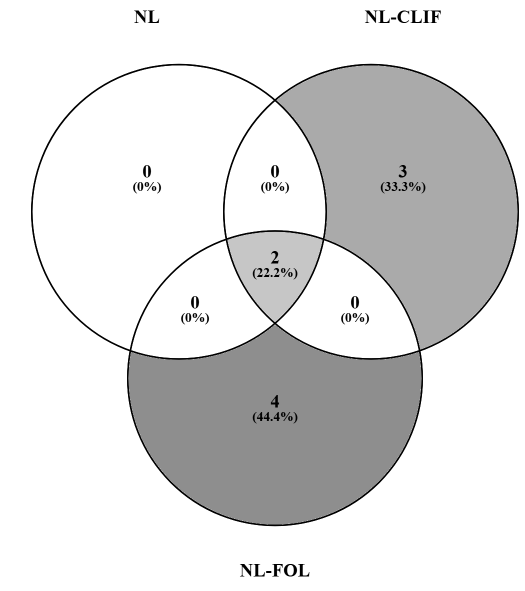}
  \caption {Left: False Positive common errors of the best FOLIO-SEMEVAL Flan-T5-large model notations. Right: False Negative common errors of the best FOLIO-SEMEVAL Flan-T5-large models.}
  \label{fig:folio_semeval_common_error_analysis}
\end{figure*}

We present a sample of qualitative error analysis for our best SEMEVAL and FOLIO-SEMEVAL models. Figure \ref{fig:semeval_common_error_analysis} shows the count of common errors among the 3 best SEMEVAL models and Figure \ref{fig:folio_semeval_common_error_analysis} mirrors the analysis for the best 3 FOLIO-SEMEVAL models. 

Table \ref{tab:semeval_qualitative_error_analysis} shows some common and uncommon errors made by the best 3 SEMEVAL models. Table \ref{tab:folio_semeval_qualitative_error_analysis} shows the same analysis for the best 3 FOLIO-SEMEVAL models.

\begin{table*}
  \centering
  \resizebox{\linewidth}{!}{
  \begin{tabular}{p{0.2\linewidth} | p{0.6\linewidth} | p{0.1\linewidth} | p{0.2\linewidth} | p{0.1\linewidth} | p{0.1\linewidth} | p{0.1\linewidth}lllllll}
    \hline
    \textbf{ID}  & \textbf{Syllogism} & \textbf{Validity} & \textbf{Plausibility} & \textbf{NL} & \textbf{NL-CLIF} & \textbf{NL-FOL} \\
    \hline
e773bd8c-fa53-4e9c-8ec6-7d978e0601ac & Every single object can fly. It is known that some boats cannot fly. It follows that some boats are not objects. & true & false &  & X &  \\
\hline
eae77932-d7db-4ce8-b5b1-6c3c951ef553 & It is the case that some pencils are white. There are some sheets of paper that are not white. This implies that some sheets of paper are not pencils. & false & true &  & X & X \\
\hline
fc1a0164-31a4-48aa-ab9d-176a66b93bfe & Anything that is a square is also a quadrilateral. No circle is a square. It follows that no circle is a quadrilateral. & false & true & X &  &  \\
\hline
6725d344-7c13-4d44-a97d-a4a1b89f858d & The category of celestial bodies and the category of planets are mutually exclusive. Not a single planet is a star. It must be the case that a portion of stars are not celestial bodies. & false & false & X & X &  \\
\hline
fb637f9c-1c26-4302-9c01-94c061bd352c & The category of chairs and the category of living things do not overlap. The group of living things and the group of inanimate objects are mutually exclusive. A portion of inanimate objects are not chairs. & false & true & X & X & X \\
\hline
fc53be1a-0b75-4995-bcb2-bc3a878a0cb2 & "No animal that is a house pet is a feline. Not a single cat is a house pet. This demonstrates that some cats are not felines. & false & false & X &  & X \\
\hline
  \end{tabular}
  }
  \caption{SEMEVAL sample qualitative error analysis by model: X indicates the model made a mistake in the prediction.}
  \label{tab:semeval_qualitative_error_analysis}
\end{table*}

\begin{table*}
  \centering
  \resizebox{\linewidth}{!}{
  \begin{tabular}{p{0.2\linewidth} | p{0.6\linewidth} | p{0.1\linewidth} | p{0.2\linewidth} | p{0.1\linewidth} | p{0.1\linewidth} | p{0.1\linewidth}lllllll}
    \hline
    \textbf{ID}  & \textbf{Syllogism} & \textbf{Validity} & \textbf{Plausibility} & \textbf{NL} & \textbf{NL-CLIF} & \textbf{NL-FOL} \\
    \hline
6725d344-7c13-4d44-a97d-a4a1b89f858d & The category of celestial bodies and the category of planets are mutually exclusive. Not a single planet is a star. It must be the case that a portion of stars are not celestial bodies. & false & false & X & X & X \\
\hline
45d4df27-8269-4ecb-9ead-f6571561f3d5 & There is at least one spoon that is a kitchen tool. Some of the utensils are spoons. Consequently, some of the utensils are kitchen tools. & false & true & X & X & X \\
\hline
fc1a0164-31a4-48aa-ab9d-176a66b93bfe & Anything that is a square is also a quadrilateral. No circle is a square. It follows that no circle is a quadrilateral. & false & true & X &  & X \\
\hline
fb637f9c-1c26-4302-9c01-94c061bd352c & The category of chairs and the category of living things do not overlap. The group of living things and the group of inanimate objects are mutually exclusive. A portion of inanimate objects are not chairs. & false & true &  & X & X \\
\hline
e773bd8c-fa53-4e9c-8ec6-7d978e0601ac & Every single object can fly. It is known that some boats cannot fly. It follows that some boats are not objects. & true & false &  & X &  \\
\hline
c62f852d-16fd-4eda-b380-b85d9a17f9e2 & Every single creature is a mammal. It is true that no mammal is an amphibian. Therefore, it's the case that no amphibian is a creature. & true & false &  &  & X \\
\hline
  \end{tabular}
  }
  \caption{FOLIO-SEMEVAL sample qualitative error analysis by model: X indicates the model made a mistake in the prediction.}
  \label{tab:folio_semeval_qualitative_error_analysis}
\end{table*}

\end{document}